\documentclass[journal, letterpaper]{IEEEtran}

\usepackage{graphicx}

\usepackage{xcolor}
\usepackage{hyperref}

\hypersetup{
    colorlinks=true,   
    linkcolor=blue,    
    citecolor=blue,    
    filecolor=blue,    
    urlcolor=blue      
}

\usepackage{url}

\usepackage{amsmath}

\usepackage{textgreek}	
\usepackage{listings}
\usepackage{csvsimple}
\usepackage{longtable}
\usepackage{url}

\begin{document}

	\title{Proprioceptive Origami Manipulator}

\author{Aida Parvaresh$^{1}$, Arman Goshtasbi$^{1}$, Jonathan Tirado$^{1}$, Ahmad Rafsanjani$^{1*}$
\thanks{$^{1}$SDU Soft Robotics, Biorobotics Section, The Maersk Mc-Kinney Moller Institute, University of Southern Denmark (SDU), 5230 Odense M, Denmark
      {\tt\small (email: \{aidap, argo, joti, ahra\}@sdu.dk)}\newline$^{*}$Author to whom correspondence should be addressed.} 
}
	\markboth{}{}
	\maketitle
    
\begin{abstract}
Origami offers a versatile framework for designing morphable structures and soft robots by exploiting the geometry of folds. Tubular origami structures can act as continuum manipulators that balance flexibility and strength. However, precise control of such manipulators often requires reliance on vision-based systems that limit their application in complex and cluttered environments. Here, we propose a proprioceptive tendon-driven origami manipulator without compromising its flexibility. Using conductive threads as actuating tendons, we multiplex them with proprioceptive sensing capabilities. The change in the active length of the tendons is reflected in their effective resistance, which can be measured with a simple circuit. We correlated the change in the resistance to the lengths of the tendons. We input this information into a forward kinematic model to reconstruct the manipulator configuration and end-effector position. This platform provides a foundation for the closed-loop control of continuum origami manipulators while preserving their inherent flexibility.
\end{abstract}


\section{Introduction}

Continuum manipulators possess infinite passive degrees of freedom, allowing them to continuously deform and surpass traditional rigid robotic arms in flexibility and adaptability~\cite{russo2023continuum}. Among the various actuation methods employed to drive continuum manipulators, tendon-driven mechanisms are the most widely used approach. In this method, forces are transmitted from motors through tendons that run along the manipulator's structure, ensuring reliable stability and manipulability~\cite{amanov2021tendon}.

Current tendon-driven continuum manipulators often rely on an inextensible, elastic, slender backbone decorated with evenly spaced spacer disks~\cite{rao2021model,parvaresh2022dynamics}. The incompressibility of the backbone in these traditional continuum manipulators limits their ability to adjust their effective length. At the same time, their low flexural stiffness makes them susceptible to vibrations during dynamic movements.
In recent years, new backbone designs have emerged to overcome these limitations and enhance the motion capabilities of tendon-driven continuum robots. For example, spring-based backbones allowed for both extension and compression~\cite{li2018design}. Variable stiffness mechanisms, such as granular jamming~\cite{choi2021tendon} and pneumatic chambers~\cite{wockenfuss2022design}, could change the rigidity of the backbone.
Telescopic magnetic backbones enabled adjusting the arm length using magnetic spacer disks with alternating pole orientation~\cite{nguyen2015tendon}.

\begin{figure}[tb]
\centering
\includegraphics[width=\columnwidth]{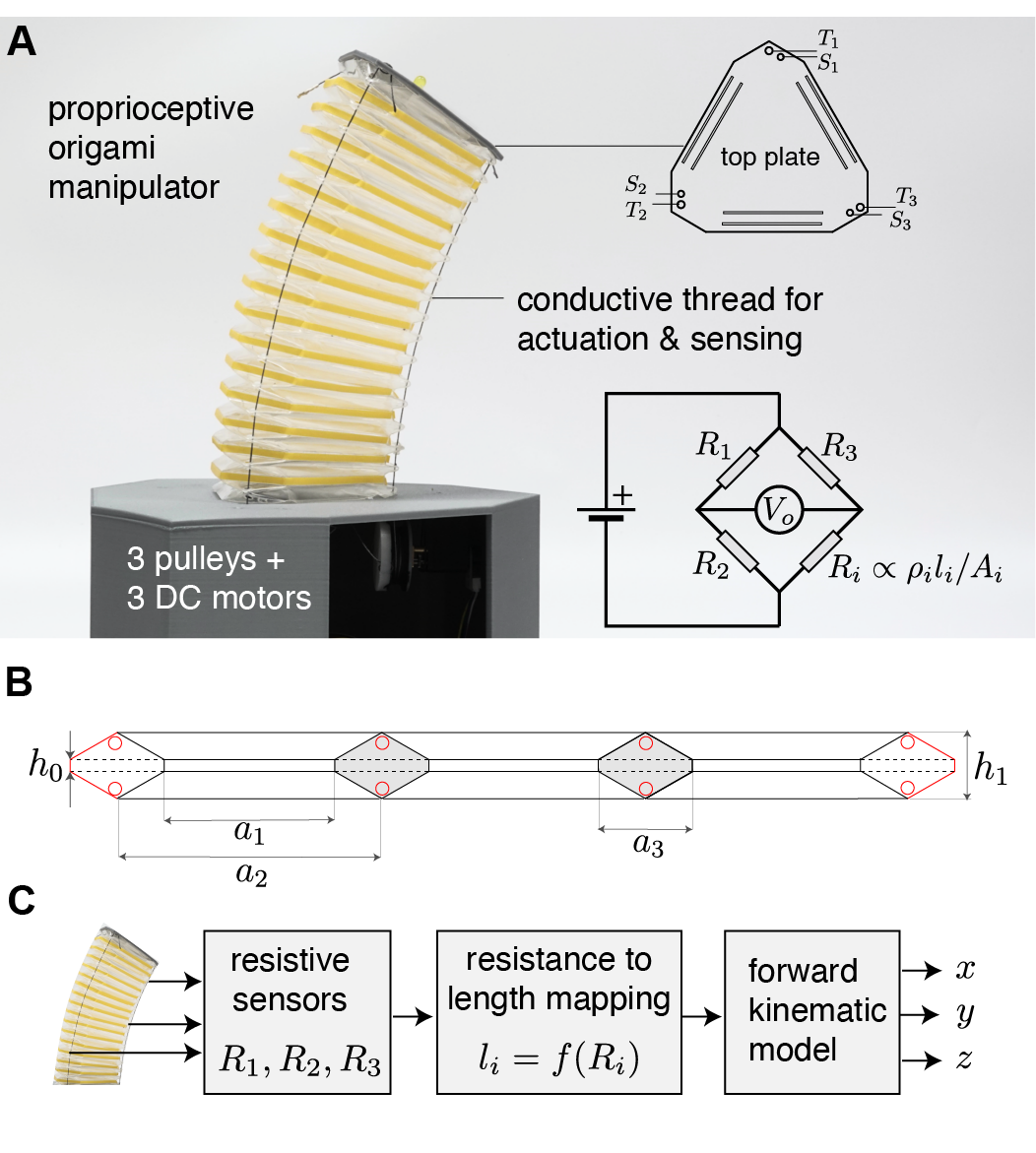}
\caption{\textbf{A}. The continuum origami manipulator with integrated proprioceptive sensing. Variations in the conductive thread's active length alter its effective resistance, which is measured via a Wheatstone bridge to infer the actuator shape. \textbf{B}. Fold pattern of the origami manipulator. \textbf{C}. The workflow for reconstructing the end effector position from sensor readings. }
\label{Fig1}
\end{figure}

\textit{Origami} technique enabled cost-effective and quick construction of 3D tubular structures by introducing patterned folds into thin flat sheets \cite{santoso2021origami}, offering lightweight, compressible, and extensible backbone designs for continuum manipulators with programmable flexural stiffness~\cite{son20224d}. 
The shape, mechanical properties, and even dynamic behavior of the origami structures can be programmed by controlling the geometric parameters of crease patterns~\cite{rus2018design,zhai2018origami}.
Several researchers adopted origami-inspired designs to improve the versatility and adaptability of tendon-driven manipulators and crawling robots~\cite{junfeng2024modular,yoo2023design,santoso2021origami,zhang2022design,zhang2023bioinspired,zhang2024origami, parvaresh2024metamaterial}. 
These designs have demonstrated benefits in creating self-assembling, deployable, adaptable, and impact-resistant robots~\cite{morgan2016approach}. However, they often lack integrated sensing, which is crucial for enabling real-time feedback, precise control, and enhanced adaptability in dynamic environments. 

Integrating traditional solid-state rigid sensors often compromises the flexibility of origami manipulators. This has led researchers to only use encoders (in tendon-driven systems) or pressure sensors (in pneumatic systems) for low-level feedback, while others rely solely on forward modeling assumptions to estimate the robot's position without sensing. 
Also, external motion capture systems were employed for proprioceptive sensing feedback in origami manipulator~\cite{santoso2021origami}.
However, the dependence of optical systems on reflective markers, lighting conditions, and manipulator position negatively affects the sensing, and reliance on an external setup limits their autonomous operation in various environments.
Alternatively, other researchers used a mounted force sensor on an origami end-effector to measure interaction forces with the environment and achieve positioning by regulating stiffness without extensive sensory information~\cite{zhang2022design, zhang2023bioinspired}.

This work proposes an origami manipulator with integrated proprioceptive sensing (see Fig.~\ref{Fig1}). Conductive threads serve as tendons in the tendon-driven actuation system, powering the manipulator and functioning as resistive sensors, with each tendon's effective resistance proportional to its active length. The manipulator combines high torsional stiffness with omnidirectional bending and predictable contraction. We characterized the performance of the proprioceptive sensing mechanism under different actuation protocols and fed the data into a forward kinematic model to estimate the manipulator's trajectory (see Fig.~\ref{Fig1}C). We demonstrated the approach's effectiveness by comparing these results with those from a motion capture system. The embodied proprioception enhances origami manipulators by enabling shape inference from sensor readings without compromising flexibility.

\begin{figure*}[t]
\centering
\includegraphics[width=\textwidth]{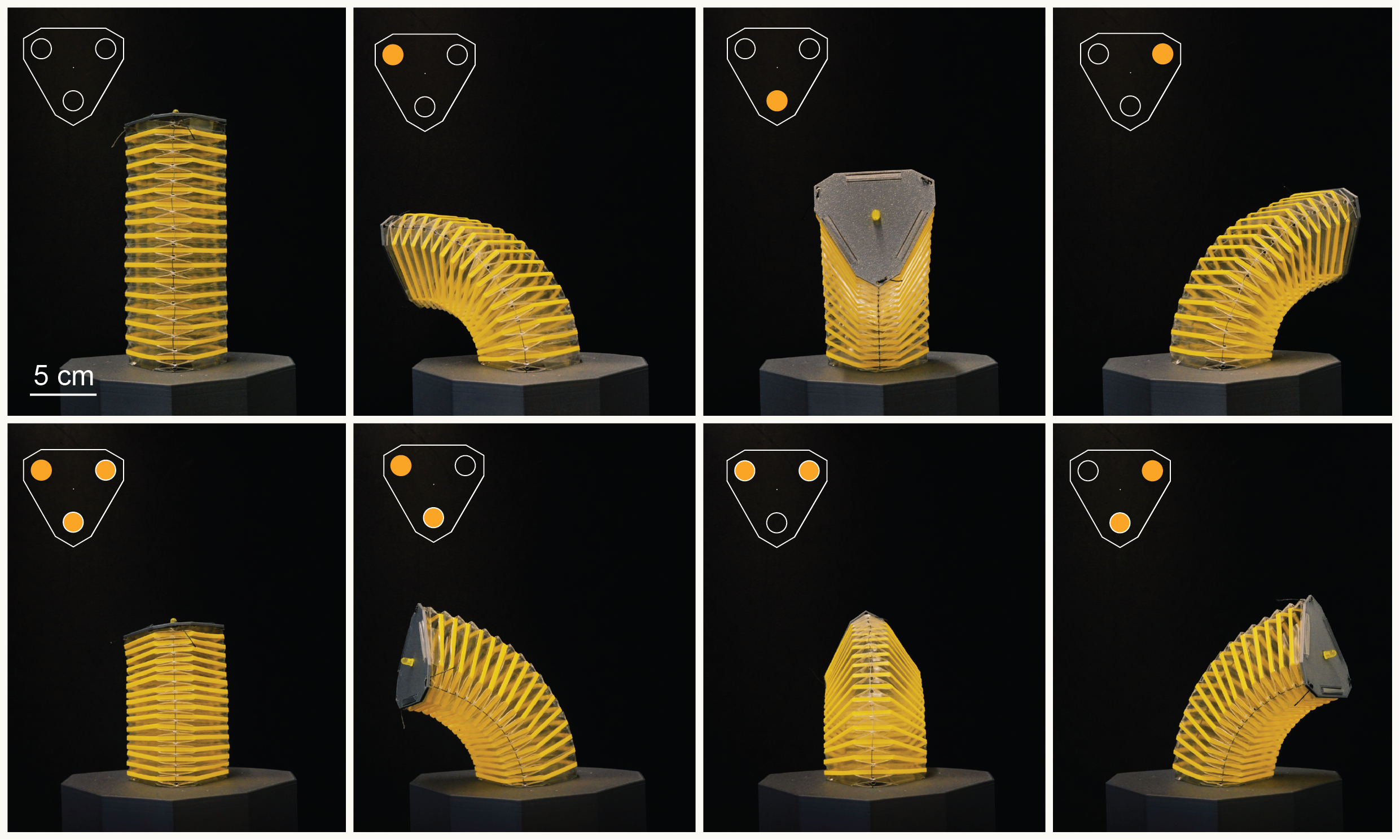}
\caption{Omnidirectional movements of the continuum origami manipulator when different combinations of three tendons are pulled. All active tendons, highlighted in yellow on top-view schematics, are pulled to the same length. }
\label{Fig2}
\end{figure*}

\section{Materials and Methods}
\label{sec:background}

\subsection{Design and Fabrication}
In nature, the elephant trunk is a unique example of combining flexibility and strength for powerful manipulation. The trunk of African elephant \textit{Loxodonta africana} uses wrinkles and folds to modulate stiffness, providing a valuable source of inspiration for soft robotics~\cite{schulz2022skin,schulz2024elephants}.
Following these biological principles, we adopted the \textit{Yoshimura} origami pattern to create a continuum manipulator that, like an elephant trunk, is flexible while resilient under applied forces to overcome the limitation of soft robotic manipulators with limited load-bearing capacity.
The original Yoshimura structure allows for multi-directional bending and axial deformation while also providing resistance to torsion~\cite{santoso2021origami}.
To enhance overall stability, load-bearing capacity, resistance to external loads, and prevent collapsing, we introduced prismatic rings with flat surfaces between axial units, which limit contraction when the structure is fully compressed. 

The origami manipulator is made of Polyethylene terephthalate (PET) sheets (0.1 mm thick), which were chosen for their excellent foldability, high strength-to-weight ratio, and low cost. The origami creases were engraved using a laser cutting machine (VLS/ULS 2.30 Universal Laser Systems) at 80\% speed and 9\% power for the folds, while the boundaries were cut at 22\% speed and 9\% power.
Then, the engraved sheet was pre-folded along the crease pattern, embossing the fold lines. The sheet was then rolled into a bellow shape, with the edges connected using double-sided adhesive (3M, 200 MP, acrylic, 0.127 mm thickness). An additional layer of triangular patches was added to the corners to balance the joint region and ensure symmetric deformation. The origami manipulator was further folded into its final shape. 
The repeating unit cell of the origami manipulator and corresponding geometrical parameters are illustrated in Fig.~\ref{Fig1}B, where red lines indicate cuts, black dashed lines represent valley folds, and black solid lines depict mountain folds. This fundamental unit is characterized by $a_1=50$\;mm, $a_2=78$\;mm, $a_3=28$\;mm, $h_0=3.5$\;mm, $h_1=19.5$\;mm.

\subsection{Tendon-driven Actuation}
The origami manipulator was actuated using a tendon-driven actuation system comprised of three tendons made of three-ply conductive thread (stainless steel, 0.5 mm diameter) and DC motors (Pololu 298:1 Micro Metal Gearmotor) with pulleys. Tendons were routed through holes at the corners of the origami structure and tied to the top plate of the manipulator. Each tendon is attached to custom-designed pulleys mounted on DC motors housed in a box at the base (see Fig.~\ref{Fig1}A).
By utilizing differential actuation of the tendons, the origami structure can assume a variety of configurations, allowing the manipulator to perform complex bending and contraction motions, as demonstrated in Fig.~\ref{Fig2}.

\subsection{Embodied Proprioception}
To sense the manipulator's configuration, we used a stainless steel fiber conductive thread (three-ply, 0.5 mm diameter, 7 $\Omega$) as a tendon.
Therefore, the same tendons used for actuation act as proprioceptive sensors. 
The resistance of the conductive thread is proportional to its length, described by $R = {\rho l}/{A}$, where $\rho$ represents the material resistivity, $l$ is the thread length, and $A$ is the cross-sectional area. Rolling the i-th tendon shortens its active length, and therefore, its effective resistance $R_i$ decreases, which can be simply measured using a Wheatstone Bridge (WB) circuit. 
The bridge is arranged with two series-parallel branches of resistors connected between a voltage supply and ground, producing zero voltage difference between the two branches when balanced. Two known resistors, \( R_1 \) and \( R_2 \), are placed in one branch, while another known resistor, \( R_3 \), is paired with the unknown resistor \( R_x \) in the opposite branch (see Fig.~\ref{Fig1}).

\begin{equation}
    R_i = R_4 \times \frac{1 - \left( \frac{R_2}{R_1 + R_2} - \frac{V_{\text{out}}}{V_{\text{in}}} \right)}{\frac{R_2}{R_1 + R_2} - \frac{V_{\text{out}}}{V_{\text{in}}}}, \quad \text{for } i = 1, 2, 3
\end{equation}

We recorded $V_{\text{out}}$ through the analog pin of an Arduino board connected to a personal computer running MATLAB for real-time sensing and post-processing, where the noise was filtered, and the resistance change was correlated to the change in tendon's active length.

\subsection{Forward Kinematics}

We adopted a forward kinematics model based on the Piecewise Constant Curvature (PCC) assumption~\cite{hannan2003kinematics} while accommodating both bending and change in the length of the manipulator in the absence of torsional motion.
The manipulator's backbone is approximated by a series of $n$ rigid links connected through rotary and prismatic joints, representing the robot's bending and contraction motions~\cite{keyvanara2023geometric}. The forward kinematics of the single-segment robot follows:
\begin{equation}
    \begin{aligned}
    \begin{bmatrix}
    X_{E}\\
    Y_{E}\\
    Z_{E}\\
    \end{bmatrix}  = \frac{L+\Delta L}{n}
    \begin{bmatrix} \cos\phi\sum_{j=1}^{n}\sin\frac{(2j-1)\theta}{2n}\\
    \sin\phi\sum_{j=1}^{n}\sin\frac{(2j-1)\theta}{2n}\\
    \sum_{j=1}^{n}\cos\frac{(2j-1)\theta}{2n}\\
    \end{bmatrix} 
    \end{aligned}
    \label{eq:multi_3D_ff}
\end{equation}
where $[X_E, Y_E, Z_E]$ represents the end-effector position, $L$ denotes the segment length, $\Delta L$ is the change in the segment length, $\theta$ indicated the bending angle of the segment, $\phi$ is the deflection angle relative to the $X$ axis. 
Additionally, we used the following kinematic relations to convert the joint space to configuration space, as detailed in \cite{KinematicWalker}. 

\begin{equation}
    \begin{aligned}
    & L = \frac{l_1+l_2+l_3}{3}\\
    & \phi = \tan^{-1}{\frac{\sqrt{3}}{3}\frac{l_3+l_2-2l_1}{l_2-l_3}}\\
    & \theta = \frac{2\sqrt{l_1^2+l_2^2+l_3^2-l_1l_2-l_2l_3-l_1l_3}}{3d}
    \end{aligned}
    \label{eq:joint2config}
\end{equation}
where $l_i$ is the length of the $i-$th tendon, with tendon $1$ positioned along the y-axis, and $d$ represents the distance between the center and the tendon. 

\section{Results and Discussion}

\subsection{Mechanical Characterization}

\begin{figure}[tb]
\centering
\includegraphics[width=\columnwidth]{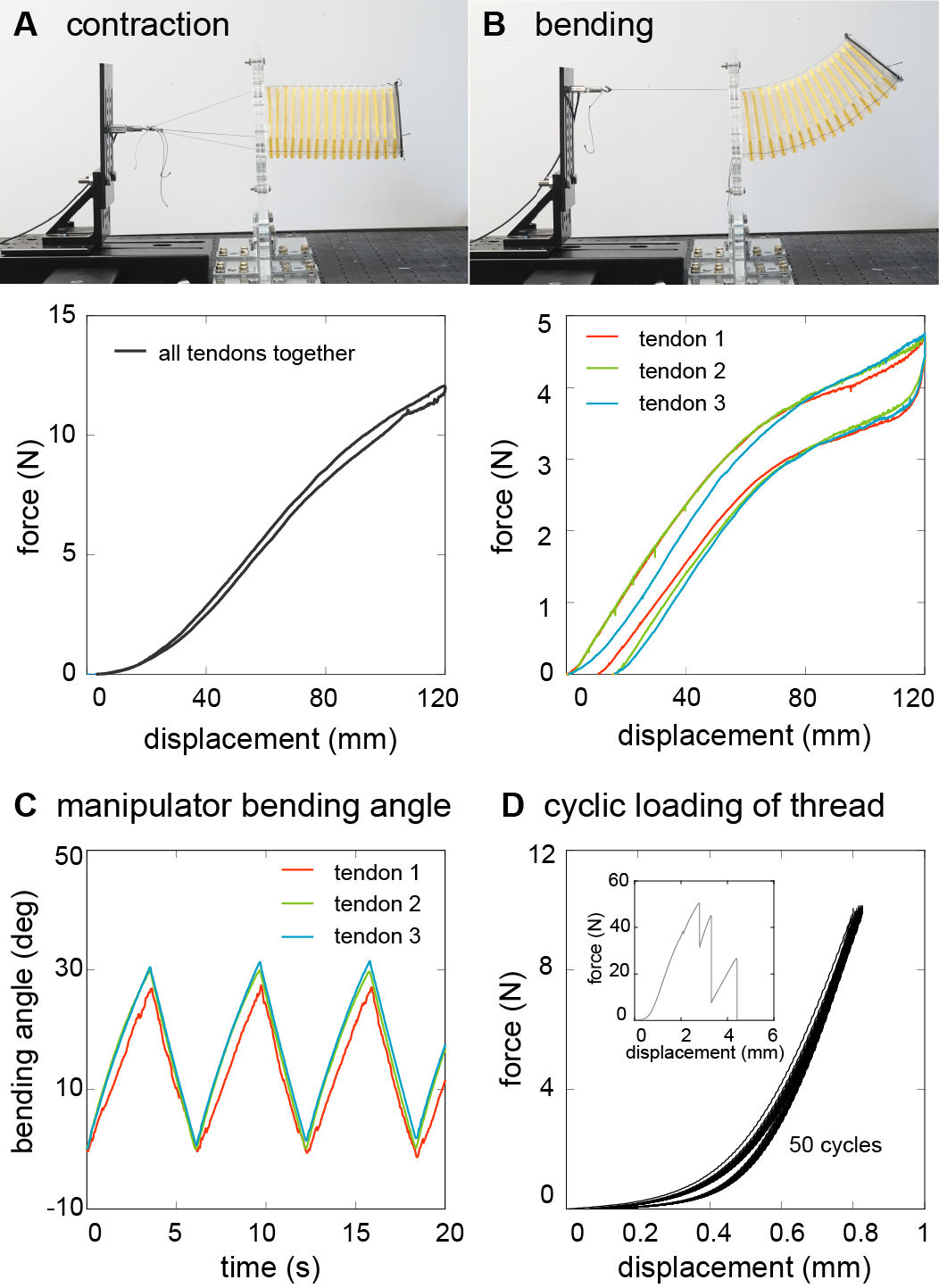}
\caption{\textbf{A}. Characterization of the axial deformation of the origami body. \textbf{B}. Characterization of the bending deformation of the origami body.  \textbf{C}. Characterization of the bending angle for each tendon.  \textbf{D}. Characterization of conductive thread under cyclic loading. The inset shows the thread strength under tensile test.}
\label{Fig3}
\end{figure}

\textit{Axial and bending stiffness}:
To understand the relationship between the force applied by the tendons and the axial and bending behavior of the origami manipulator, we designed a setup to simulate the tendon tensions. A motorized linear stage (Thorlabs LTS300) was used to pull each tendon at a constant speed of 5 mm/s until complete compression/ side bending of the structure was achieved.
The linear stage was equipped with a 5 lb load cell (Futek LSB200) and an analog amplifier (Futek IAA100) to record the data accurately. Each experiment was repeated three times for each tendon, ensuring reliable results. 
The results for uniform contraction (all tendons pulled simultaneously) and bending (each tendon pulled individually) are shown in Fig.~\ref{Fig3}A and~\ref{Fig3}B, respectively. 
These measurements indicate that the maximum force the tendons apply for axial contraction is approximately 13 N. Notably, the manipulator can withstand higher compression forces (up to 80 N) when the flat rings come into contact.
Additionally, the maximum force applied to each tendon during bending ranges from 4.5 to 5 N, providing valuable insight for selecting an appropriate motor for actuation.
Fig.~\ref{Fig3}C demonstrates the unidirectional bending of the origami manipulator when each tendon is actuated individually. The bending angle was characterized by tracking the orientation of the edge of the top plate. Minor discrepancies between tendons are attributed to variations in the fabrication process.

\textit{Thread strength}:
To determine the mechanical properties of the conductive thread, tensile tests were conducted using a universal testing machine (EZ Test, Shimadzu) following the ASTM D3822 standard for textile fibers. The sample length was 5 cm, and it was placed between a fixed gripper and a loading gripper. The test was performed at the crosshead speed of 10 mm/min to prevent quick breaking. The results are presented in the inset of Fig.~\ref{Fig3}D, showing the initial fiber breakage occurs at a tensile force of 38 N, and it continues until the failure of all fibers.
For further study, we applied cyclic tension to the conductive thread up to 10 N (approximately twice the maximum force applied to each tendon) for 50 cycles, which confirmed its durability under cyclic loading as demonstrated in Fig.~\ref{Fig3}D. 

\begin{figure}[h]
\centering
\includegraphics[width=\columnwidth]{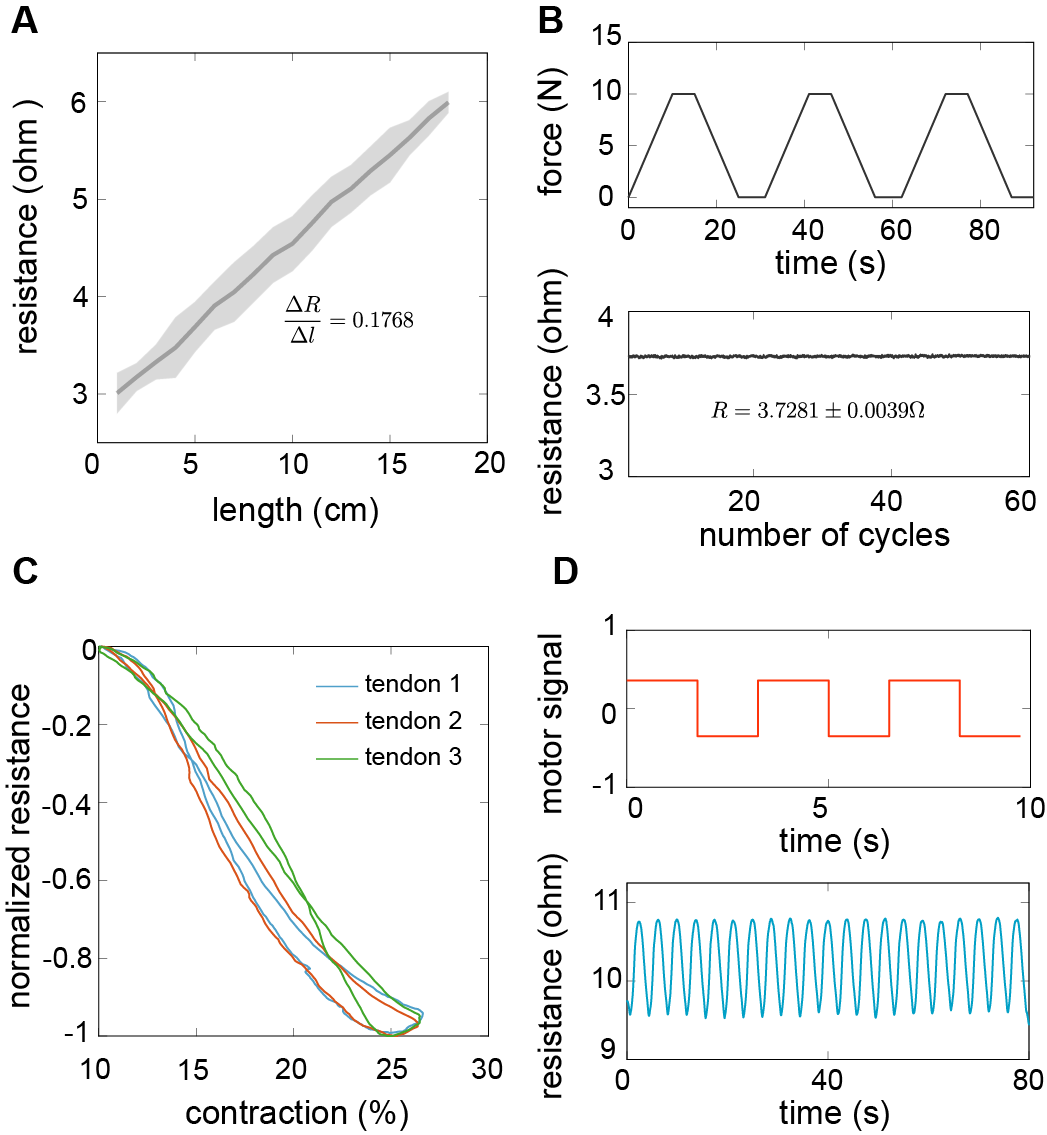}
\caption{\textbf{A}. Characterization of the resistance of the conductive thread. \textbf{B}. Top: Profile of the cyclic force applied to the thread using a tensile testing machine. Bottom: Corresponding resistance measurements over multiple cycles. \textbf{C}. Normalized resistance variation of each tendon as a function of the contraction of each edge of the origami manipulator.  \textbf{D}. Top: Profile of cyclic motor activation. Bottom: Repeatability of sensor measurements for Tendon 2 under cyclic motor activation.}
\label{Fig4}
\end{figure}

\subsection{Sensing Calibration and Validation}
To study the relationship between the length and resistance of the conductive thread, we measured the resistance of an 18 cm long thread in the free state (not in the manipulator) at several points with an LCR meter (GW Instek, model 6100) and repeated the measurements three times. The results presented in Fig.~\ref{Fig4}A show a linear correlation between the length and resistance of the thread. 
To study the effect of tensile force on resistance measurement, we placed the conductive thread between the grippers of the tensile test machine and attached the LCR meter clips to both ends. Then, we applied cyclic loading up to 10 N (top figure in Fig.~\ref{Fig4}B). The results showed an insignificant change in resistance measurement ($ R=3.7281\pm 0.0039 \Omega $) under applied force. This confirms that resistance sensor can be reliably used without concerns about the influence of actuation forces on resistance measurement.  

The resistance of the conductive threads for each motor was measured by the designed Wheatstone Bridge (WB) using an Arduino board, and the data was transmitted through serial communication to MATLAB for further calculations and post-processing. To correlate the measured resistance with the variation in tendon length, which is an important feature in the closed-loop control schemes, we first collected the resistance data, and at the same time, we recorded the variation in tendon lengths with image processing for each tendon.
For this purpose, we placed 10 markers along the manipulator near the tendons, collected their position through the camera, and analyzed them in object tracking software (Tracker software \cite{Brown2010}).  
Then, we calculated the distance between consecutive segments to determine changes in length and contraction of the manipulator. For a bending up to $30^\circ$, the resistance of tendons 1, 2, and 3 varies in the range $9.65-10.2 \; \Omega$, $9.5-10.5 \;\Omega$, $9.65-10.35\;\Omega$, respectively. To develop a unified model for all tendons, we normalized the resistance variations of each tendon. 
The plots of normalized resistance variation versus contraction in the manipulator edge is reported in Fig~\ref{Fig4}C.
We mapped the measured resistance to the tendon length using a third-order polynomial function, $  l_i=f(R_i)= a + b {R_i} + c {R_i}^2 + d {R_i}^3$, where $l_i$ is the predicted tendon length and $R_i$ is the measured resistance and the curve parameters are $a=0.102$, $b=-0.172$, $c=-0.205$, and $d=-0.173$  ($R^2=0.97$, $RMSE=1.25$ mm).

Repeatability is a key performance parameter that reflects a sensor's ability to consistently produce the same readings under identical conditions. For our resistive sensor, several factors can affect repeatability, including mechanical wear on the thread, weak connections between the thread and sensor pin, and variations in initial conditions. To mitigate these effects, a calibration process is recommended. In this study, we evaluated the sensor's repeatability over continuous operation cycles. Fig.~\ref{Fig4}D shows the sensor's output over 20 bending cycles when tendon 1 was actuated, demonstrating the consistency of our measurements.

To validate the sensor's performance, we tested its readings for cyclic bending with an increasing amplitude, where the bending angle increased by 5$^\circ$ in each cycle ($0^\circ-20^\circ-0^\circ-25^\circ-0^\circ-30^\circ-0^\circ-35^\circ$). We collected the resistance measurements from the second tendon and reconstructed its length using the mapping (third-order polynomial equation) extracted for correlating the resistance and length values. The left plot in Fig.~\ref{Fig5}A illustrates the recorded sensor's measurements, while the right plot compares the tracked length variations (obtained by analyzing the captured video) with the reconstructed length variation derived from sensing to length mapping. The results indicate that the sensor can accurately track the actuator's length changes and its performance under varied conditions highlights the sensor's robustness and ability to reliably capture length variations.

\subsection{Path Reconstruction via Proprioception}

\begin{figure}[tb]
\centering
\includegraphics[width=\columnwidth]{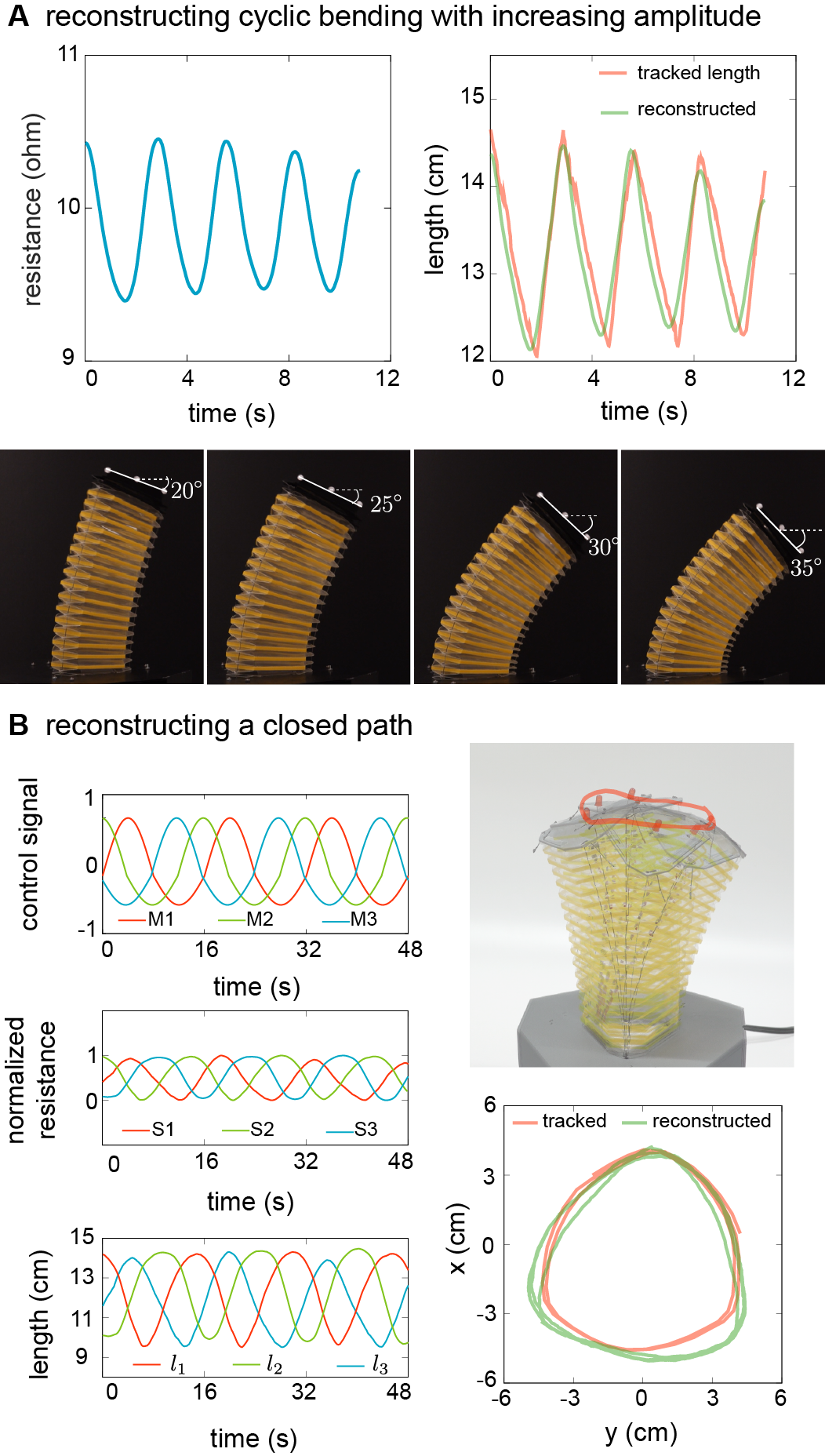}
\caption{\textbf{A}. Reconstructing tendon length during cyclic bending with increasing amplitude. Left: measured resistance, Right: tracked vs constructed length.  \textbf{B}. Reconstruction of a closed path using the resistance sensor readings paired with the forward kinematic model. }
\label{Fig5}
\end{figure}

Finally, we used the sensing data in conjunction with the forward kinematic model introduced earlier to demonstrate the proprioceptive capability of our origami manipulator and reconstruction of the end-effector path through sensing measurements.
The manipulator was actuated using a cyclic actuation protocol with the frequency of $f=0.0625$ Hz and phase shift of $2\pi/3$ between the actuation signal of the motors to follow a closed trajectory (see top-left plot in Fig.~\ref{Fig5}B). During this process, resistance values were measured (middle-left plot), and the actuator lengths were reconstructed using the resistance to length mapping model for the sensors (bottom-left plot). These reconstructed actuator lengths were then fed into the forward kinematic model to calculate the coordinates of the end effector. The resulting values were compared with the real-world measurements obtained from a motion capture system (OptiTrack), as illustrated in bottom-right plot Fig.~\ref{Fig5}B. The reconstructed path shows a reasonable agreement with the tracked positions (see Supporting Movie).

\section{Conclusion}
In this study, we developed a tendon-driven origami manipulator that maintains flexibility while enhancing strength—an often neglected aspect in soft robotics. To address this challenge, we integrated flat facets between the folds of the origami structure, significantly improving both axial and flexural stiffness. This design modification stabilizes the manipulator during dynamic motions and reduces vibrations.
We integrated proprioception into the manipulator by utilizing the same actuating tendons as resistive sensors. The changes in the active length of these conductive threads correspond to variations in their effective resistance, which can be easily measured with a Wheatstone bridge. By correlating these resistance changes to the tendon lengths, we input this information into a forward kinematic model to reconstruct the manipulator's configuration. This method provides a robust foundation for closed-loop control of continuum origami manipulators while preserving their inherent flexibility. 

However, certain challenges must be addressed before proceeding further. The primary issue is improving the hardware setup to achieve higher sensing resolution. Since the range of resistance changes is small ($1-2 \;\Omega$), designing a circuit to amplify the signal before post-processing and filtering would enhance accuracy. Additionally, the fabrication and assembly process affects sensor readings across different tendons. To minimize errors, it is crucial to maintain consistency in pretension levels, sensor connections, and initial lengths. In the next steps, integrating the proposed embodied sensing mechanism with an inverse kinematic framework would enable closed-loop control. Expanding the current design to multi-module origami manipulators and incorporating exteroceptive effects using the same sensing mechanism would enhance the system's adaptability in complex environments, such as handling payloads and interacting with contacts.

\section{Acknowledgement}
This work was supported by the Villum Young Investigator grant 37499.

\appendix
The supporting video of this article can be found at this link: \url{https://youtu.be/z7LMPFLUEuI}.

\bibliographystyle{IEEEtrans}
\bibliography{reference} 

\end{document}